\documentclass[letterpaper, 10 pt, conference]{ieeeconf}  

\IEEEoverridecommandlockouts                              

\usepackage{graphics} 
\usepackage{epsfig} 
\usepackage{mathptmx} 
\usepackage{times} 
\usepackage{amsmath} 
\usepackage{amssymb}  

\newtheorem{proposition}{Proposition}

\newtheorem{remark}{Remark}

\newcommand{\norm}[1]{\lVert #1 \rVert}
\newcommand{\ee}{\mathrm{ee}}
\newcommand{\obj}{\mathrm{obj}}
\newcommand{\vobj}{v_{\mathrm{obj}}}
\newcommand{\R}{\mathbb{R}}

\usepackage[caption=false,font=footnotesize]{subfig}
\usepackage{graphics} 
\usepackage{epsfig} 
\usepackage{mathptmx} 
\usepackage{times} 
\usepackage{amsmath} 
\usepackage{mathtools}
\usepackage{amssymb}  
\usepackage{xcolor}
\usepackage{graphicx}
\usepackage{subcaption} 
\usepackage{booktabs}
\usepackage{siunitx}
\usepackage{xcolor}
\usepackage{subcaption}
\usepackage{dsfont}
 \usepackage{float} 
\title{\LARGE \bf
Deep Reinforcement Learning for Robotic Manipulation under Distribution Shift with Bounded Extremum Seeking
}

\title{\LARGE \bf
Planning for Change: Reinforcement Learning Combined with Bounded Extremum Seeking for Robotic Control under Distribution Shift
}

\author{Shaifalee Saxena$^{1, 2}$, Rafael Fierro$^{1}$ and Alexander Scheinker$^{2}$
\thanks{This work was supported by the U.S. Department of Energy (DOE),
Office of Science, Office of High Energy Physics contract number
9233218CNA000001 and the Los Alamos National Laboratory LDRD
Program Directed Research (DR) project 20220074DR. 
R. Fierro’s work is sponsored by Air Force Research Laboratory (AFRL) under agreements FA9453-18-2-0022 and FA9550-22-1-0093.}
\thanks{$^{1}$Shaifalee Saxena and Rafael Fierro are with Department of Electrical and Computer Engineering, University of New Mexico, Albuquerque, NM 87106, USA,
        {\tt\small \{shaifalisaxena, rfierro\}@unm.edu}}%
\thanks{$^{2}$Shaifalee Saxena and Alexander Scheinker are with Los Alamos National Lab, Los Alamos, NM 87547, USA
        {\tt\small \{shaifalees, ascheink\}@lanl.gov}}%
}
\begin{document}

\maketitle
\thispagestyle{empty}
\pagestyle{empty}

\begin{abstract}

Reinforcement learning has shown strong performance in robotic manipulation, but learned policies often degrade in performance when test conditions differ from the training distribution. This limitation is especially important for reliable robot planning and control in contact-rich tasks such as pushing and pick-and-place, where changes in goals, contact conditions, or robot dynamics can drive the system out-of-distribution at inference time. In this paper, we investigate a hybrid controller that combines reinforcement learning with bounded extremum seeking (ES) to improve robustness under such conditions. In the proposed approach, deep deterministic policy gradient (DDPG) policies are trained under standard conditions on the robotic pushing and pick-and-place tasks, and are then combined with bounded ES during deployment. The RL policy provides fast manipulation behavior, while bounded ES ensures robustness of the overall controller to time variations when operating conditions depart from those seen during training. The resulting controller is evaluated under several out-of-distribution settings, including time-varying goals and spatially varying friction patches. Crucially, under reasonable assumptions, we prove finite time RL-based convergence of a robotic arm to an object after which bounded ES achieves finite-time convergence to a bounded velocity moving target. We provide analytic formulas for both convergence times, and an ability to tune the convergence times by our choice of gains.

\end{abstract}

\section{Introduction}

Planning, the ability to make purposeful, goal-directed decisions and to carry them out reliably as conditions evolve, is a cornerstone of long-term robot autonomy \cite{meli2023planning}. Since learned policies alone struggle to sustain this reliability under distribution shift, hybrid architectures that couple AI-based decision making with control-theoretic structure at execution time have emerged as a promising route to resilient behavior \cite{emam2021safe}, a paradigm the RL-extremum-seeking controller studied here instantiates for contact-rich manipulation.

Robotic manipulation tasks such as pushing and pick-and-place are fundamental examples of contact-rich decision making, since they involve intermittent contact, hybrid dynamics, and tight perception-to-action coupling \cite{billard2019trends}. These characteristics make them difficult to model and control using purely analytic methods, especially when the robot must reason over long-horizon interactions such as reaching, contacting, grasping, and transporting an object. Reinforcement learning has therefore emerged as a natural framework for such tasks, since it can learn complex feedback strategies directly from interaction without requiring an accurate closed-form model of contact \cite{nguyen2019review}. However, the same properties that make manipulation attractive for RL also make it highly sensitive to changes in the task and environment, particularly when the policy is executed outside the training regime. 

A substantial body of work has shown that RL can succeed on robotic manipulation tasks under fixed training conditions. Hindsight Experience Replay \cite{andrychowicz2017hindsight} and the multi-goal Fetch benchmark suite made sparse-reward pushing and pick-and-place substantially more tractable and established these tasks as common testbeds for algorithmic comparison \cite{plappert2018multi}. Beyond benchmark settings, RL has also been used to learn pushing-grasping synergies \cite{zeng2018learning} and to improve picking through interaction-driven exploration \cite{deng2019deep}. These results demonstrate that learned policies can acquire effective contact-rich manipulation behaviors, but they primarily reflect performance within the training distribution or under carefully designed benchmark assumptions.

 Robotic manipulation is often partially observable and context dependent, with uncertainty arising from noisy sensing, unknown object properties, and changing environments \cite{lauri2022partially}. RL policies can overfit to their training distribution and degrade under shifted test conditions \cite{cobbe2019quantifying}. Methods such as RL$^2$ \cite{duan2016rl}, Model-Agnostic Meta-Learning \cite{finn2017maml}, and model-based transfer learning \cite{cho2024model} all seek faster adaptation to changing conditions. Contextual RL further models environment variation through a set of contexts and seeks policies that generalize across such variations \cite{benjamins2021carl}. Large-scale procedural training with aggressive domain randomization has also shown strong generalization gains in other domains, such as locomotion \cite{liu2025locoformer}. However, for contact-rich manipulation, training over such broad distributions can be computationally demanding, and it is generally infeasible to cover every possible deployment scenario during training, which motivates inference-time approaches to improve robustness under unseen deployment conditions.

Another approach to improve robustness is to combine reinforcement learning with structured feedback or adaptive control. In such architectures, the RL policy is paired with a controller that provides nominal regulation, adaptation, or safety. In residual RL, a conventional feedback controller is augmented with an RL-learned residual policy to handle contact- and friction-rich robotic manipulation tasks \cite{johannink2019residual}. Related ideas appear in approaches that combine RL with impedance control \cite{kulkarni2022learning}, model reference adaptive control \cite{guha2021online}, model predictive control \cite{romero2025actor}, and robust control barrier functions \cite{emam2021safe}. These works suggest that robustness can be pursued not only through broader training distributions, but also through more structured feedback architectures at deployment. In our setting, however, robustness depends on the ability to adapt online once contact-rich interaction begins, since the resulting dynamics may vary during execution. This motivates the use of Extremum Seeking (ES) as the adaptive component in our hybrid architecture.

Extremum Seeking has been applied in robotics to uncertain motion control \cite{koropouli2016extremum} and optimized path tracking \cite{bajpai2024investigating}. Bounded ES is well suited in our setting because it offers guaranteed bounds on control efforts and parameter update rates despite acting on noisy, analytically unknown time-varying systems \cite{scheinker2013model}, and has been studied for a wide range of systems including ones with non-differentiable and discontinuous controllers \cite{scheinker2016bounded}, and has been applied for various physical systems \cite{xie2005autopilot, scheinker2021extremum, de2022event, saxena2026improved, saxena2026mahalanobis}. This motivates the present study, where RL is used to learn the manipulation behavior and bounded ES is used at inference time to adapt that behavior when operating conditions differ from those seen during training. 

In this paper, we investigate a combined RL and bounded ES controller for robotic pushing and pick-and-place. Deep deterministic policy gradient (DDPG) policies are first trained on standard Fetch manipulation tasks, and bounded ES is then combined with the learned controller during inference. The objective is to leverage the complementary strengths of both methods: RL uses data to learn fast manipulation behavior when the operating regime remains close to the training distribution, while bounded ES provides robust model-independent feedback when time-varying goals drive the system away from that regime. In particular, RL is used to provide rapid task entry, including approach, contact acquisition, and, for pick-and-place, grasp formation, which are difficult for standalone ES to discover reliably from scalar performance feedback alone. In pushing, standalone ES fails to identify a useful initial pushing direction and in pick-and-place, it likewise fails to reliably discover the reaching and grasp-formation behavior needed to establish grasp and initiate transport. We evaluate the combined controller under several out-of-distribution settings, including time-varying goals and spatially varying friction patches without fine-tuning RL at inference time.
\section{Background}
%
\subsection{Bounded Extremum Seeking for Time-Varying Systems}
\label{ES}
%
We briefly review the bounded extremum seeking (ES) results most relevant to this work; see \cite{scheinker2016bounded, scheinker2014extremum} for full developments. Let $x\in\mathbb{R}^n$ evolve according to
\begin{equation}
	\dot{x} = f(x,t) + g(x,t)u(x,t),
	\label{dxdt}
\end{equation}
where $f:\mathbb{R}^{n}\times\mathbb{R}\rightarrow\mathbb{R}^n$ and $g:\mathbb{R}^{n}\times\mathbb{R}\rightarrow\mathbb{R}^{n\times m}$ are unknown, and $u:\mathbb{R}^{n}\times\mathbb{R}\rightarrow\mathbb{R}^m$ is the control input. We focus on two special cases of Eq. \eqref{dxdt} that motivate the use of bounded ES in this paper.

The first is stabilization. In this case, let $g:\mathbb{R}^{n}\times\mathbb{R}\rightarrow\mathbb{R}^{n}$ and let $u:\mathbb{R}^{n}\times\mathbb{R}\rightarrow\mathbb{R}$ be scalar. Choosing the Lyapunov candidate $V(x)=x^T x$, define the feedback law
\begin{equation}
	u(x,t) = \sqrt{\alpha\omega}\cos(\omega t + k V(x)).
\end{equation}
For sufficiently large $\omega$, the system can be approximated by the weak limit-averaged dynamics
\begin{equation}
	\dot{\overline{x}} = f(\overline{x},t) -\frac{k\alpha}{2} g(\overline{x},t)g^T(\overline{x},t)\nabla_{\overline{x}}V(\overline{x}).
\end{equation}
The averaged control direction is therefore given by the positive semidefinite matrix
$g(\overline{x},t)g^T(\overline{x},t)\geq 0$, which removes the sign ambiguity of the original control direction. As a result, the origin can be stabilized by selecting a sufficiently large gain $k\alpha>0$.

A second case, which is especially relevant for optimization, is obtained by taking $f(x,t)=0$, letting $g(x,t)$ be diagonal with entries $g_i(x,t)$, $i\in\{1,\dots,n\}$, and using a vector control input $u$. Suppose that we observe a noisy measurement,
\begin{equation}
	y(x,t) = J(x,t) + n(t),
\end{equation}
of an analytically unknown, time-varying cost function $J(x,t)$ that we wish to minimize. Then the system becomes
\begin{equation}
	\dot{x}_i = g_i(x,t)u_i(x,t), \quad y(x,t) =J(x,t)+n(t).
\end{equation}
For this setting, the bounded ES feedback law is chosen as
\begin{equation}
	u_i(x,t) = \sqrt{\alpha\omega_i}\cos(\omega_i t + k y(x,t)), 
	\quad \omega_i = r_i\omega, \quad r_i \neq r_j,
	\label{uESopt}
\end{equation}
where the perturbation frequencies are pairwise distinct. The corresponding averaged dynamics satisfy
\begin{equation}
	\dot{\overline{x}} = -\frac{k\alpha}{2} g(\overline{x},t)g^T(\overline{x},t)\nabla_{\overline{x}}J(\overline{x},t) = -\frac{k\alpha}{2} g_i^2(\overline{x},t)\frac{\partial J(\overline{x},t)}{\partial \overline{x}_i}, \label{ES_ave}
\end{equation}
a gradient descent on the unknown objective $J(\overline{x},t)$.

For both cases above, \cite{scheinker2016bounded} shows that for any compact set $K\subset\mathbb{R}^n$, any horizon $T>0$, and any $\delta>0$, there exists $\omega^*$ such that for all $\omega>\omega^*$,
\begin{equation}
	\| x(t) - \overline{x}(t) \| < \delta, \qquad \forall t\in [0,T]. \label{eq:ES_conv}
\end{equation}
Moreover, this guarantee can be extended to $T\to\infty$ when $\overline{x}(t)$ converges to a stable equilibrium. These results make bounded ES a useful model-free tool for optimizing noisy, analytically unknown, time-varying objective functions and for stabilizing unknown time-varying systems. In this paper, we use these properties to complement deep reinforcement learning (DRL)-based feedback controllers in robotic manipulation under changing deployment conditions.

%
\subsection{Deep Reinforcement Learning for Feedback Control}
\label{sec:DDPG}
%

We model the control problem as a discounted Markov decision process
$M=(S,A,p,r,\gamma)$ with discount factor $\gamma\in[0,1)$. At time $t$, the agent observes a state $s_t\in S$, applies an action $a_t\in A$, receives reward $r_t=r(s_t,a_t)$, and transitions to a new state $s_{t+1}\sim p(\cdot\,|\,s_t,a_t)$. The objective is to learn a policy $\pi$ that maximizes the discounted return
\begin{equation}
J(\pi)=\mathbb{E}_\pi\!\Big[\sum_{t=0}^{\infty}\gamma^t r_t\Big].
\end{equation}

For continuous control, deterministic actor-critic methods are widely used because they avoid the high variance associated with sampling over stochastic actions. The deterministic policy gradient (DPG) theorem \cite{silver2014deterministic} states that for a differentiable deterministic policy $\mu_\theta:S\to A$,
\begin{equation}
\label{eq:dpg}
\nabla_\theta J(\mu_\theta)=
\mathbb{E}_{s\sim \rho^{\mu_\theta}}
\!\left[\nabla_\theta \mu_\theta(s)\;\nabla_a Q^{\mu_\theta}(s,a)\big|_{a=\mu_\theta(s)}\right],
\end{equation}
where $\rho^{\mu_\theta}$ denotes the discounted state visitation distribution and $Q^{\mu_\theta}(s,a)$ is the corresponding action-value function.

DDPG implements Eq. \eqref{eq:dpg} using deep neural networks together with two stabilizing mechanisms inherited from deep Q-learning: experience replay for off-policy updates and slowly moving target networks for both the actor and critic \cite{lillicrap2020continuous}. Let $\mu_{\theta^\mu}$ denote the actor and $Q_{\theta^Q}$ the critic, with target networks $\mu_{\theta^{\mu'}}$ and $Q_{\theta^{Q'}}$. These targets are updated by Polyak averaging,
\begin{equation}
\theta^{\mu'}\leftarrow\tau\theta^\mu+(1-\tau)\theta^{\mu'}, 
\qquad
\theta^{Q'}\leftarrow\tau\theta^{Q}+(1-\tau)\theta^{Q'},
\end{equation}
for $0<\tau\ll 1$. During data collection, exploratory actions are generated as
\begin{equation}
a_t=\mu_{\theta^\mu}(s_t)+\varepsilon_t.
\end{equation}

\paragraph{Critic update}
Given a minibatch $\{(s_i,a_i,r_i,s'_i,d_i)\}_{i=1}^N$ sampled from the replay buffer $\mathcal D$, where $s'_i$ is the next state and $d_i\in\{0,1\}$ is the terminal indicator, the temporal-difference target is
\begin{equation}
y_i = r_i + \gamma (1-d_i)\,
Q_{\theta^{Q'}}\!\big(s'_i,\,\mu_{\theta^{\mu'}}(s'_i)\big),
\end{equation}
and the critic parameters are updated by minimizing
\begin{equation}
\mathcal{L}(\theta^Q) = \tfrac{1}{N}\sum_{i=1}^N
\big(Q_{\theta^Q}(s_i,a_i)-y_i\big)^2 .
\end{equation}

\paragraph{Actor update}
Using samples from replay, the deterministic policy gradient is approximated by
\begin{equation}
\label{eq:actor-grad}
\nabla_{\theta^\mu} J \;\approx\;
\tfrac{1}{N}\sum_{i=1}^N
\Big[\nabla_a Q_{\theta^Q}(s,a)\big|_{a=\mu_{\theta^\mu}(s_i)}\Big]\;
\nabla_{\theta^\mu}\mu_{\theta^\mu}(s_i).
\end{equation}

\section{DRL Policy Training}
\label{sec:DRL_training}

\begin{table}[t]
\centering
\caption{\small DDPG training hyperparameters}
\label{tab:ddpg_params}

\begin{tabular*}{0.8\columnwidth}{@{\extracolsep{\fill}}ll@{}}
\toprule
\textbf{Hyperparameter} & \textbf{Value} \\
\midrule
State dimension $n_s$ & 28 \\
Action dimension $n_a$ & 4 \\
Discount factor $\gamma$ & 0.99 \\
Actor learning rate & $1\times10^{-4}$ \\
Critic learning rate & $1\times10^{-4}$ \\
Replay buffer size $|\mathcal{D}|$ & $10^6$ transitions \\
Batch size $B$ & 256 \\
Soft target update $\tau$ & 0.005 \\
Exploration noise & $\mathcal{N}(0, 0.1)$ \\
\bottomrule
\end{tabular*}
\end{table}

\begin{figure}[t] 
\centering 
\includegraphics[width=1.0\columnwidth]{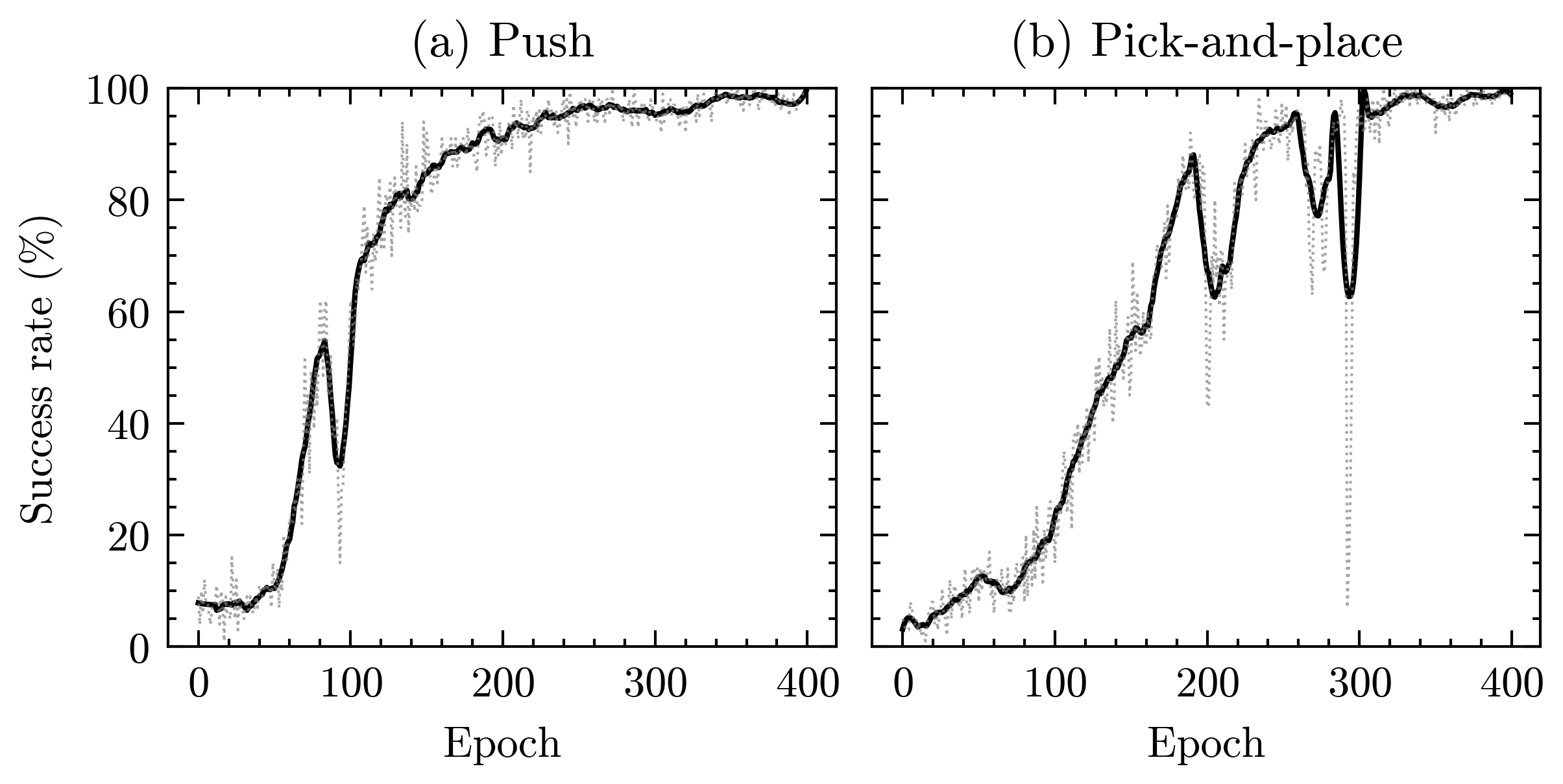} 
\caption{DDPG training success rates for Fetch push and pick-and-place. The dotted gray curves denote the raw success rates, while the solid black curves denote the smoothed training trends. Each training epoch consists of 100 episodes.} 
\label{fig:fetch_sr} 
\vspace{-0.6cm}
\end{figure}

The manipulation policies are trained with DDPG. The DDPG formulation and update laws were summarized in Section~\ref{sec:DDPG}. We use the FetchPush and FetchPickAndPlace environments from the Fetch benchmark suite \cite{plappert2018multi}, implemented in Gymnasium-Robotics and simulated using the MuJoCo physics engine \cite{todorov2012mujoco}. These environments are based on a 7-DoF Fetch Mobile Manipulator with a two-fingered parallel gripper. For these environments, training is carried out in a goal-conditioned setting: at the beginning of each episode, both the initial object pose and the desired goal are randomized within the workspace. This prevents the policy from overfitting to a single target configuration and instead encourages learning a goal-conditioned policy that maps the current state and desired goal to an action. 
%
\subsection{Task, Observation, and Action Representation}
%
Two robotic manipulation tasks are considered: Fetch push and Fetch pick-and-place. In both tasks, the objective is to drive the manipulated object to a specified goal position within the episode horizon.

In the \textit{push} task, the object remains on the tabletop and the robot must establish contact and push it toward the desired planar goal. In the \textit{pick-and-place} task, the robot must reach the object, grasp it, lift it, and transport it to the desired three-dimensional goal position.

At each time step, the simulator returns an observation vector together with the desired goal. We define the RL state as $s_t = [o_t, g_t]^T \in \mathbb{R}^{28}$, where $o_t \in \mathbb{R}^{25}$ is the environment observation and $g_t \in \mathbb{R}^{3}$ is the desired goal position. The observation $o_t$ contains the robot end-effector position, object position, relative object-gripper displacement, finger states, object orientation, object linear and angular velocities, gripper linear velocity, and finger velocities. 

The action is a continuous-valued 4-dimensional control input defined as:
$\begin{bmatrix} \Delta x_t & \Delta y_t & \Delta z_t & a^{gr}_t \end{bmatrix}^{\top} \in [-1,1]^4,$
where the first three components command Cartesian end-effector motion and the fourth component commands gripper opening or closing.
%
\subsection{Reward Design}
\label{subsec:reward}
%

We use a dense shaped reward to encourage both approaching the object and transporting it to the goal. Let $p_t^{\mathrm{ee}} \in \mathbb{R}^3$, $p_t^{\mathrm{obj}} \in \mathbb{R}^3$, and $g_t \in \mathbb{R}^3$ denote the end-effector, object, and goal positions, respectively. Define
\begin{equation}
d_1 = \|p_t^{\mathrm{ee}} - p_t^{\mathrm{obj}}\|_2, \qquad
d_2 = \|p_t^{\mathrm{obj}} - g_t\|_2, \label{eq:distances}
\end{equation}
and the reward as
\begin{equation}
r_t = -d_1 - d_2 + 2\,\mathds{1}_{\!\left\{d_2 \le \delta\right\}},
\label{eq:reward}
\end{equation}
where $\delta = 0.05$ and $\mathds{1}_{\{d_2 \le \delta\}}$ denotes the indicator function. Here, $d_1$ encourages the robot to reach the object and establish contact, while $d_2$ encourages the object to move toward the goal. A terminal bonus of $+2$ is provided when the object reaches the goal within the tolerance $\delta$, which helps the agent learn task completion during training. An episode is declared successful when the manipulated object lies within $0.05$~m of the desired goal position.
%


%
\subsection{Training Procedure}
%
The actor and critic are fully connected networks with two hidden layers of 256 neurons each, using Layer Normalization and ReLU activations. The actor uses a hyperbolic tangent output to produce bounded actions, while the critic takes the concatenated state-action vector $[s_t^\top\; a_t^\top]^\top$ as input and produces a scalar $Q$-value through a linear output layer.

All training settings used in this work are summarized in Table~\ref{tab:ddpg_params}. During training, mini-batches are sampled from the replay buffer, the critic is updated from the temporal-difference target, and the actor is updated through the deterministic policy gradient. The success-rate curves in Fig.~\ref{fig:fetch_sr} show that the learned policies attain high success rates on both Fetch tasks within the training distribution.

\section{ES-DRL controller}
\label{sec:ood}

During DDPG training, we used a goal-conditioned setting with initial object pose and desired goal randomized across episodes. Each episode the goal remained fixed and the table had a single frictional surface. However, performance can degrade during deployment when operating conditions at inference time differ from those encountered during training. In these settings, the largest mismatch appears after physical interaction begins, because the resulting object motion depends on local friction, contact geometry, and other effects that are not fully captured by the training distribution. To improve robustness, we use the trained RL policy for rapid approach and contact acquisition, and then hand off control to bounded extremum seeking (ES) for local online adaptation during the contact phase. The executed action is
\begin{equation}
a_t = \beta_t a_t^{\mathrm{RL}} + (1-\beta_t) a_t^{\mathrm{ES}},
\label{eq:hybrid_action}
\end{equation}
where $\beta_t \in \{0,1\}$ is generated by a supervisor.

\subsection{DRL controller}

We deploy the policy trained by DDPG. At run time we use only the frozen actor $\mu_\theta$ to prevent policy drift and the RL control command is defined as:
\(
a_t^{RL} \;=\; \mu_\theta(s_t),
\)
where $s_t$ is the current state. Exploration noise is disabled at evaluation. The critic and target networks are used only during training.

The RL controller is responsible for task entry, up to the point at which task-relevant interaction is established. For the push task, it rapidly moves the end-effector toward the block and typically discovers a useful pushing pose in far fewer steps than a purely local adaptive search. For pick-and-place, it likewise provides the reaching and grasp-formation behavior required to establish meaningful interaction with the object. This is the regime in which the history-dependent structure learned during training is most valuable.

\begin{figure}[]
    \centering
    \includegraphics[width=\linewidth]{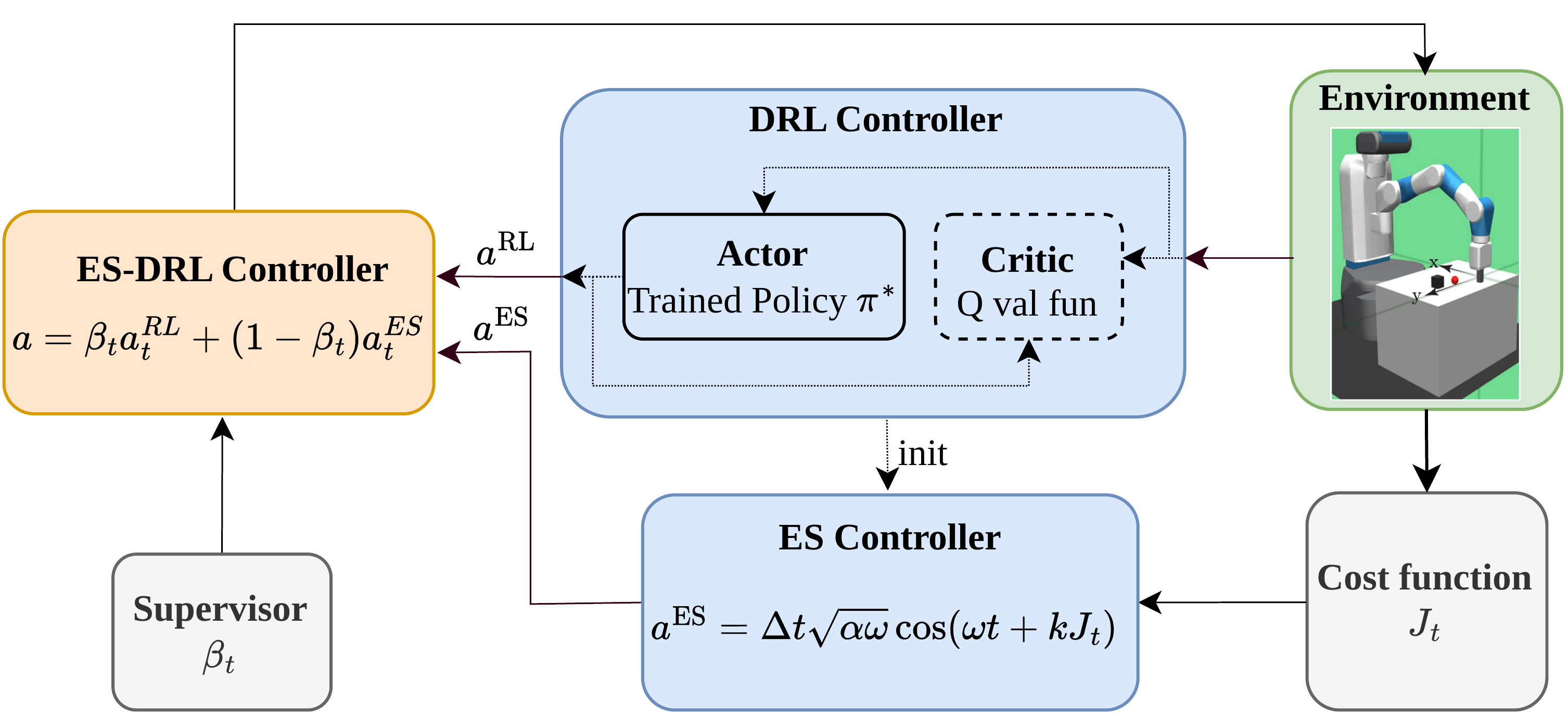}
    \caption{Architecture of the ES-DRL controller. A supervisor selects binary $\beta \in \{0,1\}$  based on the contact flag. ES is initialized from DRL (dotted).}
    \label{fig:esdrl}
    \vspace{-0.6cm}
\end{figure}

\subsection{ES controller}

Once task-relevant interaction has been established, control is transferred to bounded ES for online adaptation. Unlike the RL policy, bounded ES does not rely on a learned model of the interaction dynamics. Instead, it updates the control action from a scalar performance signal and is therefore well suited to analytically unknown, noisy, and time-varying deployment conditions \cite{scheinker2013model, scheinker2016bounded}. 

Let $n_a=4$ denote the action dimension, and let $t_c$ denote the switching instant. The ES controller is warm-started from the RL action at handoff $a_{t_c}^{\mathrm{ES}} = a_{t_c}^{\mathrm{RL}}$. 

The fourth action represents the gripper command. In the push task, it is held constant because the gripper remains closed during the entire task. For the pick-and-place task, after grasp acquisition, the gripper command provided by the RL controller is preserved during ES execution to avoid loss of grasp. Hence, bounded ES is applied only to the first three action channels, while the fourth action is kept fixed.

We adopt the discrete bounded ES update summarized in Section~\ref{ES}. For each action channel $i \in \{1,\dots,n_a-1\}$ except the last gripper-control action,
\begin{equation}
a_{i,t}^{\mathrm{ES}}
=
\Delta t \sqrt{\alpha \omega_i}
\cos\!\big(\omega_i t + k J_t\big),
\qquad i=1,\dots,n_a-1,
\label{eq:es_update}
\end{equation}

where $\alpha$ is the dither-amplitude parameter, $\omega_i$ are pairwise distinct excitation frequencies, $\Delta t$ is the ES step size, $k$ is the ES gain, and $J_t$ denotes the post-contact performance signal. Since ES is activated only after contact has been established between the gripper and the block, the reaching term is no longer dominant; consequently, the ES update is primarily driven by the object-to-goal distance $d_2$ in Eq. \eqref{eq:reward}.


\subsection{Supervisor} 

Let $\beta_t \in \{0,1\}$ denote a contact flag. For both the push and pick-and-place tasks, $\beta_t$ is initialized to 1 and switches to $0$ when the end-effector first comes into contact with the object at time $t_c$. For pick-and-place, $t_c$ is also the time when the gripper grasps the object. The switching law is:
\begin{equation}
\beta_t =
\begin{cases}
1, & t \le t_c \ \text{(RL mode)}\\
0, & t > t_c \ \text{(ES mode)}.
\end{cases}
\label{eq:beta_switch}
\end{equation}

 This switching architecture reflects the complementary strengths of the two controllers. 

%
\section{Stability and convergence analysis}
\label{sec:math}
Recall that $p_t^{\ee},\,p_t^{\obj}\in\R^3$ denote the end-effector and object positions, respectively. Let $a_t'\in\mathbb{R}^3$ denote the Cartesian components of the control action. For a control timestep $\Delta_t$, we approximate the end-effector kinematics as
\begin{equation}
    p_{t+1}^{\ee} = p_t^{ee} + \Delta_t a_t', \quad
    \frac {p_{t+1}^{\ee} - p_t^{ee}}{\Delta_t} \approx \dot p_t^{\ee} = a_t' 
\end{equation}
We assume the learned policy moves the end-effector toward the object along the line of sight, up to a bounded error, so that
\begin{equation}
\dot p_t^{\ee} = a_t^{'} = -k_{RL}\frac{p_t^{\ee}-p_t^{\obj}}{\bigl\lVert p_t^{\ee}-p_t^{\obj}\bigr\rVert} + M(s_t;\theta^\mu), \quad p_t^{\ee}\neq p_t^{\obj},
\label{eq:ee}
\end{equation}
\begin{equation}
\norm{\dot p_t^{\obj}}<\vobj, \quad \norm{M(s_t;\theta^\mu)}<C \quad\text{for a.e.\ } t\ge0 .
\label{eq:bounds}
\end{equation}
We assume that $t\mapsto p_t^{\obj}$ is absolutely continuous and obeys the velocity bound in \eqref{eq:bounds} almost everywhere and $t\mapsto M(s_t;\theta^\mu)$ is Lebesgue measurable. We also assume that our gain is sufficiently large, such that $k_{RL} > \vobj + C$, with margin $m = k_{RL}-\vobj-C >0$. To simplify notation, we define the tracking error and the Lyapunov function as
\begin{eqnarray}
    e_t &=& p_t^{\ee}-p_t^{\obj}\in\R^3, \quad r_t=\norm{e_t}, \nonumber \\
\hat e &=&\frac{e}{\norm e}\ (e\neq0), \quad V(e)=\tfrac12\norm{e}^2.
\end{eqnarray}

Wherever $e_t\neq0$, we can calculate
\begin{equation}
\dot e_t \;=\; -\,k_{RL}\,\hat e_t \;+\; w_t, \qquad w_t \;:=\; M(s_t;\theta^\mu)-\dot p_t^{\obj},
\label{eq:err}
\end{equation}
\begin{equation}
\norm{w_t}\;\le\;\norm{M(s_t;\theta^\mu)}+\norm{\dot p_t^{\obj}}\;<\;C+\vobj\;=\;k_{RL}-m,
\label{eq:wbound}
\end{equation}
by differentiating the definition of $e_t$, substituting \eqref{eq:ee}, and applying the triangle inequality $\norm{u-v}\le\norm u+\norm v$ to $w_t$, and the strict bounds \eqref{eq:bounds}.
 
\begin{remark}[Filippov solutions; set-valued map at $e=0$]\label{rem:filippov}
The right-hand side of \eqref{eq:err} is discontinuous at $e=0$ (the unit vector $\hat e$ has no limit there) and merely measurable in $t$, so solutions are understood in the Filippov sense: absolutely continuous $e(\cdot)$ with $\dot e_t\in F(e_t,t)$ for a.e.\ $t$. For fixed $t$, $F(e,t)=\bigl\{-k_{RL}\hat e+w_t\bigr\}\ \ \text{for }e\neq0$ and 
\begin{equation}
F(0,t)=\bigl\{\,w_t-k_{RL}v:\ \norm v\le1\,\bigr\},
\end{equation}
a closed ball of radius $k_{RL}$ centered at $w_t$. Since the field is bounded, $\norm{-k_{RL}\hat e+w_t}\le k_{RL}+\norm{w_t}<2k_{RL}$, and measurable in $t$, Filippov solutions exist on $[0,\infty)$ from every initial condition \cite{filippov1964differential}. So $0\in F(0,t)$ for every $t$, because $\norm{w_t}<k_{RL}$ (take $v=w_t/k_{RL}$) and exact tracking $p_t^{\ee}=p_t^{\obj}$ is an equilibrium of the regularized error dynamics for every admissible disturbance and object motion. \eqref{eq:ee} is exactly unit-vector sliding-mode control, with sliding surface $\{e=0\}$.
\end{remark}

Along any Filippov solution, at a.e.\ $t$ with $e_t\neq0$, $V(e)=\tfrac12\sum_{i}e_i^2$, so $\partial V/\partial e_i=e_i$, i.e.\ $\nabla V(e)=e$. Where $e\neq0$ the Filippov inclusion reduces to the classical equation \eqref{eq:err}, so a.e.\ there, by the chain rule,
\begin{eqnarray}
\frac{d}{dt}V(e_t) &=&\nabla V(e_t)^\top\dot e_t =e_t^\top\!\bigl(-k_{RL}\hat e_t+w_t\bigr) \nonumber \\
&=&-\frac{k_{RL}}{\norm{e_t}}\,e_t^\top e_t+e_t^\top w_t
=-\frac{k_{RL}}{\norm{e_t}}\,\norm{e_t}^2+e_t^\top w_t \nonumber \\
&=&-k_{RL}\norm{e_t}+e_t^\top w_t .
\end{eqnarray}
Therefore, for a.e.\ $t$ with $e_t\neq0$,
\begin{equation}
-\bigl(k_{RL}+\norm{w_t}\bigr)\norm{e_t} \le \frac{d}{dt}V(e_t) \le -\bigl(k_{RL}-\norm{w_t}\bigr)\norm{e_t},
\end{equation}
\begin{equation}
    \frac{d}{dt}V(e_t) \le -m\norm{e_t} = -m\sqrt{2V(e_t)}, \label{eq:pathwise}
\end{equation}
which can be shown by Cauchy--Schwarz, $|e_t^\top w_t|\le\norm{e_t}\,\norm{w_t}$, which implies
\[
-\norm{e_t}\norm{w_t}\;\le\;e_t^\top w_t\;\le\;\norm{e_t}\norm{w_t}.
\]
Based on the above, we prove our convergence result.
 
\begin{proposition}\label{prop_conv}
Let $e$ be any Filippov solution of \eqref{eq:err} with initial error $e_0=p_0^{\ee}-p_0^{\obj}$. Then we have finite-time convergence such that \quad $e_t=0$ for all
\begin{equation}
t\ge T^\ast:=\frac{\norm{e_0}}{m} =\frac{\bigl\lVert p_0^{\ee}-p_0^{\obj}\bigr\rVert}{k_{RL}-\vobj-C},
\end{equation}
and tracking after capture such that $p_t^{\ee}=p_t^{\obj}$ for all $t\ge T^\ast$: after the finite time $T^\ast$ the end-effector coincides with the moving object and remains on it.
\end{proposition}
 
\begin{proof}
Write $r_t:=\norm{e_t}$. Since $u\mapsto\norm u$ is $1$-Lipschitz and $e(\cdot)$ is absolutely continuous, $r$ is absolutely continuous; hence $r$ is differentiable a.e.\ and $r_t=r_0+\int_0^t\dot r_s\,ds$. At a.e.\ $t$ with $r_t>0$, $e$ and $r=(e^\top e)^{1/2}$ are differentiable, and
\begin{eqnarray}
\dot r_t &=& \frac{d}{dt}\bigl(e_t^\top e_t\bigr)^{1/2} = \frac{e_t^\top\dot e_t}{(e_t^\top e_t)^{1/2}}
=\frac{1}{r_t}\,\frac{d}{dt}V(e_t) \nonumber \\
&\le&  \frac{-\,m\,r_t}{r_t} =-m < 0.
\end{eqnarray}
This proves that we move towards the origin, but we also have to prove that we stay there by showing invariance of the origin. Suppose $r_{t_0}=0$ but $r_{t_1}>0$ for some $t_1>t_0$. Put $t_2:=\sup\{t\in[t_0,t_1]:\ r_t=0\}$; by continuity $r_{t_2}=0$, $t_2<t_1$, and $r>0$ on $(t_2,t_1]$. Then a.e.\ on $(t_2,t_1]$,
\begin{equation}
r_{t_1} = r_{t_2}+\int_{t_2}^{t_1}\dot r_s ds \le 0-m(t_1-t_2) < 0,
\end{equation}
contradicting $r\ge0$. Hence no Filippov solution can leave the origin once it reaches it. A continuation that sits at the origin exists, because $0\in F(0,t)$ for all $t$ by Remark~\ref{rem:filippov}; global existence of solutions is Filippov's theorem for bounded fields measurable in time. While $r>0$, integrating gives $r_t\le r_0-m\,t$. Since $r\ge0$, the error must first hit $0$ no later than $T^\ast=r_0/m$, and it remains $0$.
\end{proof}

\begin{figure*}[]
    \centering
    \includegraphics[width=0.26\textwidth]{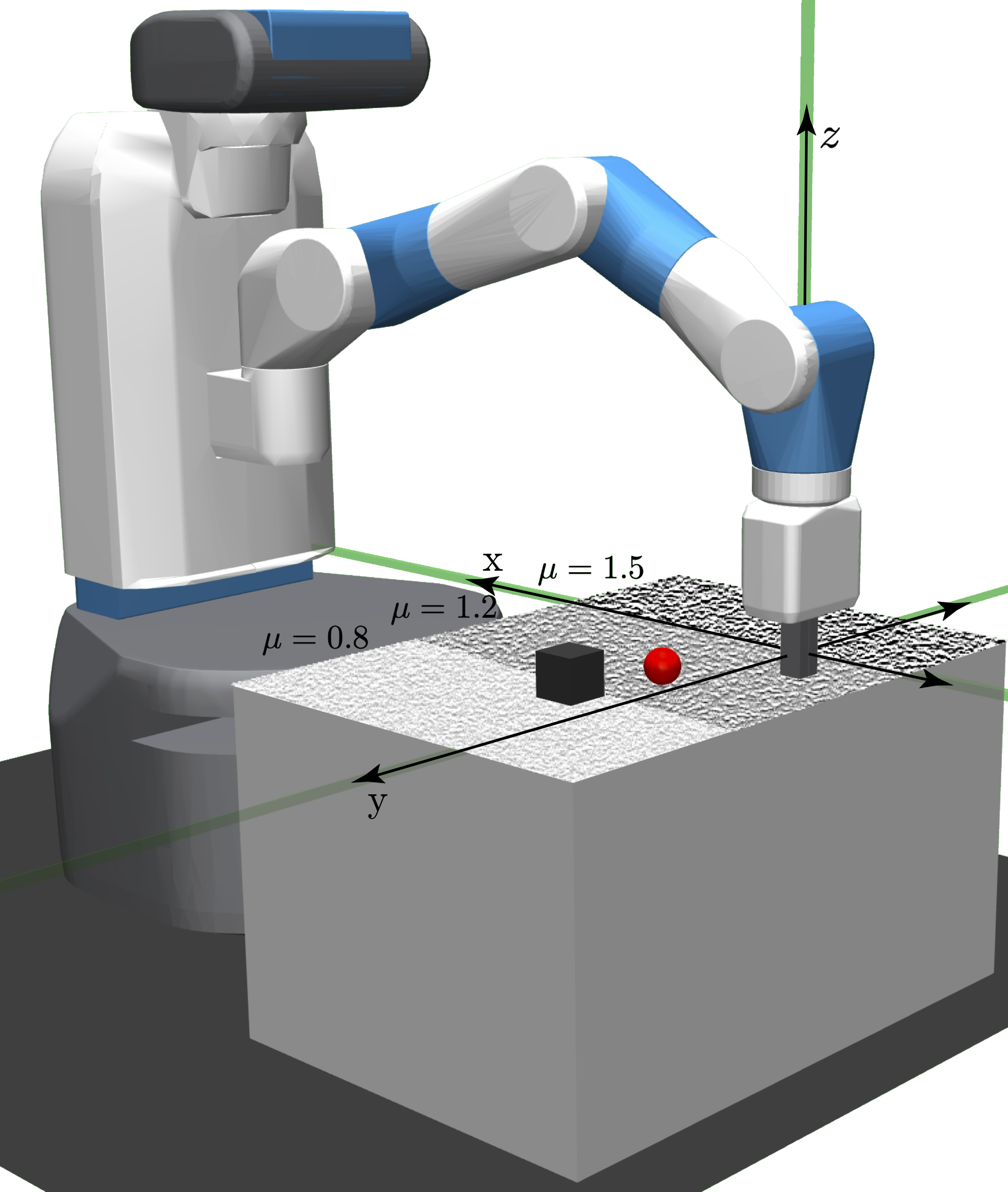}
    \includegraphics[width=0.32\textwidth]{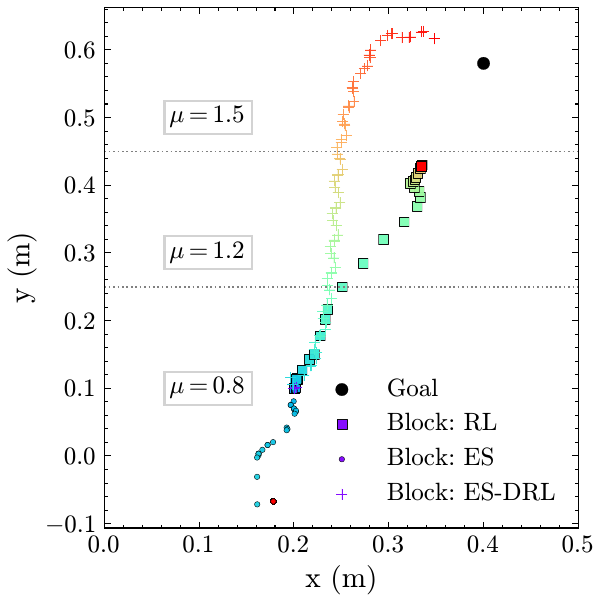}
    \includegraphics[width=0.365\textwidth]{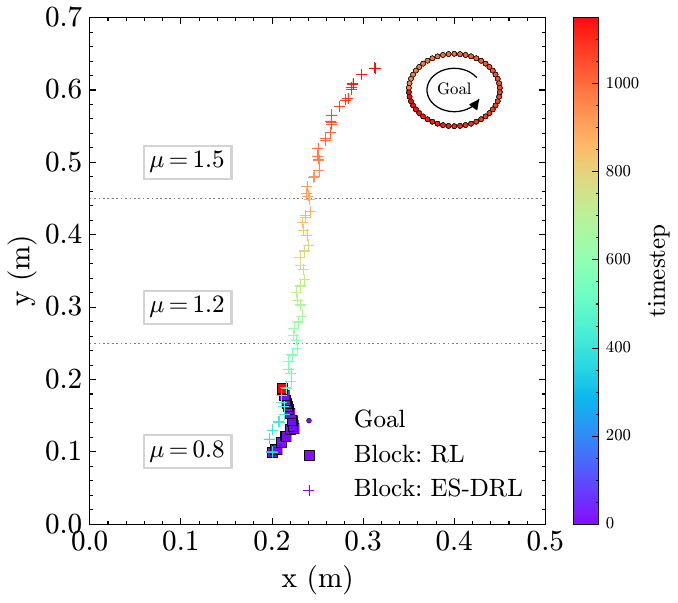}
    \caption{Robotic manipulation over spatially varying frictional surfaces. \textit{Left:} Environment with three friction patches, \(\mu = 0.8, 1.2, 1.5\). \textit{Middle:} For a fixed goal, ES lacks a good initial pushing direction, RL fails in the high-friction region, and ES-DRL drives the block toward the goal. \textit{Right:} For a time-varying goal, RL fails early, whereas ES-DRL tracks the goal.}
    \label{fig:push}
    \vspace{-0.6cm}
\end{figure*}

Therefore, for every $\delta\in(0,\norm{e_0}]$,
\[
\bigl\lVert p_t^{\ee}-p_t^{\obj}\bigr\rVert\;\le\;\delta
\qquad\text{for all }\ t\;\ge\;\frac{\norm{e_0}-\delta}{m},
\]
which is obvious from Proposition \ref{prop_conv} for such $t$, $\norm{e_0}-m\,t\le\delta$. This shows that our approach works when realistically grasping is triggered at a threshold distance $\delta$ rather than at exact coincidence. Next, we consider the push phase after the gripper establishes contact with the block at time $t_c$.
\begin{proposition}
Assuming surface friction is bounded by $M>0$ and the goal's velocity is bounded by $v_g>0$, for any $\delta>0$ there exist $\omega^*>0$ such that for all $\omega > \omega^*$, the ES-controlled object trajectory will be driven to an $\delta$-neighborhood of the goal position $g_t\in \mathbb{R}^3$ within time
\begin{equation}
    T^\ast \leq \frac{\left\| \bar{p}_{t_c}^{\mathrm{obj}} - g_{t_c} \right\|_2}{\frac{k\alpha}{2} - M - v_g},
\end{equation}
which is made arbitrarily small by increasing $k\alpha>0$.
\end{proposition}

\begin{proof} For $t>t_c$ the dynamics are described by Eq. \eqref{ES_ave}, where in the push task, $J(p^{\mathrm{obj}}_t,g_t) = \left \|p^{\mathrm{obj}}_t - g_t \right \|_2$, giving average dynamics
\begin{eqnarray}
    \dot{\bar{p}}^{\mathrm{obj}}_t = f_{\mathrm{fr}}(\bar{p}_t^{\mathrm{obj}})-\frac{k\alpha}{2}\nabla_{\bar{p}_t^{\mathrm{obj}}}J\left ( \bar{p}_t^{\mathrm{obj}},g_t \right),
\end{eqnarray}
where $f_{\mathrm{fr}}(\bar{p}_t^{\mathrm{obj}})$ represents the unknown spatially varying friction force due to the position-dependent coefficient of friction. We consider $J$ itself as a Lyapunov function for the system, for which we have
\begin{eqnarray}
    \dot{J} &=& \nabla_{p^{\mathrm{obj}}_t}J^T\dot{\bar{p}}^{\mathrm{obj}}_t + \nabla_{g_t}J^T\dot{g}_t \nonumber \\
    &=& \nabla_{p^{\mathrm{obj}}_t}J^T f_{\mathrm{fr}}(\bar{p}_t^{\mathrm{obj}})-\frac{k\alpha}{2}\nabla_{p^{\mathrm{obj}}_t}J^T \nabla_{\bar{p}_t^{\mathrm{obj}}}J + \nabla_{g_t}J^T\dot{g}_t \nonumber \\
    & \leq & \left \| \nabla_{p^{\mathrm{obj}}_t}J^T\right \| \left \| f_{\mathrm{fr}}(\bar{p}_t^{\mathrm{obj}}) \right \| - \frac{k\alpha}{2} \left \| \nabla_{\bar{p}_t^{\mathrm{obj}}}J\right \|^2 \nonumber \\
    && + v_{g} \left \| \nabla_{g_t}J\right \|,
\end{eqnarray}
where $v_g$ is a bound on the speed of the goal such that $\| \dot{g}_t \|_2 < v_g$ $\forall t$. By the above calculations
\begin{equation}
    \left \| f_{\mathrm{fr}}(\bar{p}_t^{\mathrm{obj}}) \right \| + v_{g} \frac{\left \| \nabla_{g_t}J\right \|}{\left \| \nabla_{\bar{p}_t^{\mathrm{obj}}}J \right \|} \ < \ \frac{k\alpha}{2} \left \| \nabla_{\bar{p}_t^{\mathrm{obj}}}J \right \| \quad \Longrightarrow \quad \dot{V}<0. \label{eq:ES_rate_change}
\end{equation}
Because $\nabla_{g_t} J = -\nabla_{\bar{p}_t^{\mathrm{obj}}} J$, Equation (\ref{eq:ES_rate_change}) simplifies to
\begin{equation}
    \left \| f_{\mathrm{fr}}(\bar{p}_t^{\mathrm{obj}}) \right \| + v_{g} \ < \ \frac{k\alpha}{2} \left \| \nabla_{\bar{p}_t^{\mathrm{obj}}}J \right \| \quad \Longrightarrow \quad \dot{J}<0. \label{eq:ES_rate_change_v2}
\end{equation}

Thus, the ES-controlled system drives the state to the moving objective point, and the speed of convergence is controlled by the product $k\alpha>0$. In particular, for $\bar{p}^{\mathrm{obj}}_t \neq g_t$,
\begin{equation}
    \left\| \bar{p}^{\mathrm{obj}}_t - g_t \right\|_2 > 0 \quad \Longrightarrow \quad \left\| \nabla_{\bar{p}_t^{\mathrm{obj}}} J\left( \bar{p}_t^{\mathrm{obj}}, g_t \right) \right\|_2 = 1,
\end{equation}
and for a surface with friction bound $\| f_{fr}(p^{\mathrm{obj}}_t)\|<M$, it follows that for any $k\alpha > 2(M+v_g)$ we get
\begin{equation}
    \dot{J} \leq \left\| f_{\mathrm{fr}}(\bar{p}_t^{\mathrm{obj}}) \right\| + v_g - \frac{k\alpha}{2}
\leq -\left( \frac{k\alpha}{2} - M - v_g \right) =: -c < 0.
\end{equation}
The right-hand side does not depend on the state. Therefore, by the comparison lemma, $J(t)\leq J(0)-ct$ while $J>0$, so $J$ hits zero in finite time:
\begin{eqnarray}
    J(t) &\leq& J(0) - ct \quad \Longrightarrow \quad \bar{p}_t^{\mathrm{obj}} = g_t \ \ \forall\, t \geq T^\ast, \\
    T^\ast &\leq& \frac{\left\| \bar{p}_{t_c}^{\mathrm{obj}} - g_{t_c} \right\|_2}{\frac{k\alpha}{2} - M - v_g}.
\end{eqnarray}
Furthermore, by Equation (\ref{eq:ES_conv}), for any desired $\delta>0$, if we choose sufficiently large $\omega$, then the actual trajectory $p^{\mathrm{obj}}_t$ and the averaged trajectory $\bar{p}^{\mathrm{obj}}_t$ always remain within $\delta$ of each other, therefore by the triangle inequality, for all $t\geq T^\ast$
\begin{equation}
    \| p^{\mathrm{obj}}_t - g_t \| \leq \| p^{\mathrm{obj}}_t- \bar{p}^{\mathrm{obj}}_t \| + \underbrace{\| \bar{p}^{\mathrm{obj}}_t - g_t \|}_{=0 \ \forall t > T^\ast} < \delta,
\end{equation}
so the trajectory $p^{\mathrm{obj}}_t$ converges to within any desired $\delta>0$ distance of the time-varying goal $g_t$ within time $T^\ast$.
\end{proof}

That $J$ is not differentiable on the measure zero point-set $\{0\}$ is irrelevant to the above analysis as was shown in \cite{scheinker2014non}.

\section{Results}
\label{sec:results}
%

%
\subsection{Spatially Varying Friction with Time-Varying Goals}
%

Planar pushing is highly sensitive to uncertainty in support friction \cite{zhou2017fast}. Motivated by this, we evaluate the controller on a tabletop partitioned into three friction patches with kinetic friction coefficients \(\mu = 0.8, 1.2, 1.5,\) while the policy is trained only on \(\mu=1.0\). 

For the fixed-goal case, the target is placed near the far end of the table. As shown in Fig. \ref{fig:push}, the RL-only controller can initially move the block toward the desired location, but its behavior deteriorates once the block enters the highest-friction region. The change in local contact conditions alters the relationship between end-effector motion and object motion, so the learned policy no longer produces reliable pushing behavior, and the robot’s gripper eventually loses contact with the block. A standalone ES controller is also insufficient in this setting because it lacks a learned nominal pushing direction. It cannot identify an effective initial force direction, and the block is instead driven toward the table boundary and leaves the workspace. In contrast, the ES-DRL controller continues to adapt online after contact and drives the block toward the goal despite the frictional mismatch.

We next consider a time-varying goal on the same heterogeneous surface. In this experiment, the goal follows a small circular trajectory on the tabletop, centered near the table bounds $(x_{max}, y_{max})$, with radius of a 0.05 m and a period of 200 timesteps. The RL policy fails near the beginning of the rollout and quickly departs from the operating regime seen during training, indicating poor generalization to the combined shift induced by moving references and spatially varying friction. The ES-DRL controller, however, remains stable after the RL-to-ES handoff once contact between the block and the end-effector is established and it adapts its contact actions online, allowing the block trajectory to remain close to the moving goal, as shown in Fig. \ref{fig:push}. These results show that the ES-DRL controller is especially useful when the distribution shift appears after contact and cannot be corrected by the frozen RL policy alone. 
%
\subsection{3D Tracking of a Time-Varying Goal}
%



Dynamic manipulation with time-varying targets is challenging because the robot must adapt its motion in real time as the target evolves \cite{akinola2021dynamic}. The desired goal follows a circular trajectory in the $(x,y)$ plane while its height varies slowly in $z$ relative to the tabletop. The goal trajectory $(g_x,g_y,g_z)$ is:
\begin{equation}
    (x_c, \ y_c, \ z_0) + \left (r\sin(\omega_{xy} t), \ r\cos(\omega_{xy}t), \ A_z \sin(\omega_z t) \right ),
\end{equation}
where \(x_c = 0.75 x_{max}\) and \(y_c = 0.75 y_{max}\), and $x_{max}$ and $y_{max}$ denote the maximum table bounds. Here, \(z_0 = 0.45\) denotes the table height, $r = 0.15, \ A_z = 0.20$, $\omega_{xy}=2\pi/T_{xy}$, $T_{xy} = 500$, $\omega_{z}=2\pi/T_{z}$, and \(T_z = 4000\).

\begin{figure}[]
    \centering
    \includegraphics[width=0.85\columnwidth]{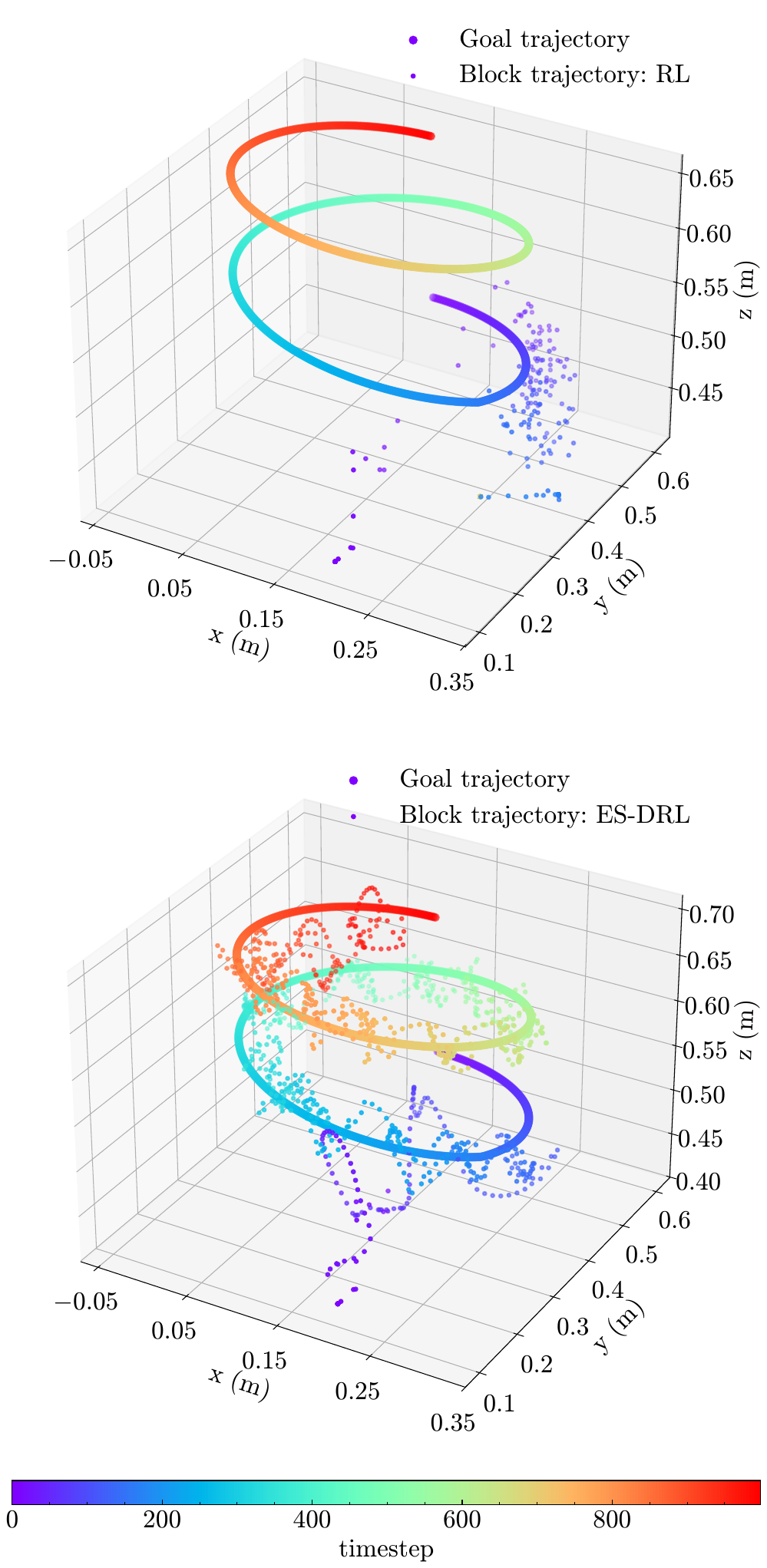}
    \caption{3D tracking of a time-varying goal. The RL-only controller (top) fails to track the goal, whereas ES-DRL (bottom) maintains substantially closer tracking.}
    \label{fig:PickAndPlace}
    \vspace{-20pt}
\end{figure}

Fig. \ref{fig:PickAndPlace} shows that the RL-only controller is unable to track this reference once the task departs from the fixed-goal regime used in training. Although the policy can initiate the nominal manipulation sequence, it does not maintain accurate object transport as the target moves, and the block trajectory deviates substantially from the desired 3D path. By contrast, the ES-DRL controller preserves the reaching and grasping behavior learned by RL and then adapts the transport phase online through bounded ES. As a result, the manipulated object follows the time-varying goal much more closely in all three coordinates, yielding significantly better 3D tracking performance than RL alone. This experiment highlights the value of inference-time ES adaptation for long-horizon manipulation tasks in which the reference evolves after grasp acquisition.

\section{CONCLUSIONS}

We investigated a hybrid extremum-seeking and reinforcement learning controller for robotic manipulation under distribution shift. A goal-conditioned DDPG policy was used to learn pushing and pick-and-place behaviors, while bounded extremum seeking was introduced at inference time to adapt the control input online. The proposed switching architecture preserves the fast task-entry behavior of RL and improves robustness during the contact-rich phase, where mismatch between training and deployment conditions is most severe. In planar pushing over spatially varying friction, the ES-only controller is unable to find a suitable initial pushing direction and eventually drives the block off the table, while the RL-only controller degrades in the high-friction region under a fixed goal and fails early when the goal is time-varying. By contrast, ES-DRL remains stable after contact and continues adapting toward the target. While tracking a 3D time-varying goal, the RL-only policy cannot maintain accurate object transport once the reference departs from the fixed-goal training regime, whereas ES-DRL achieves substantially improved tracking. These results demonstrate that bounded ES can effectively complement learned manipulation policies at inference time, improving robustness to changing contact conditions and time-varying references without retraining. For both RL and ES we proved tunable finite-time convergence. Future work will consider a hybrid supervisory policy with repeated contact-triggered switches, allowing the controller to revert to RL for contact re-acquisition if ES loses contact and then return to ES once interaction is re-established.


\bibliographystyle{IEEEtran}
\bibliography{mybibfile}

@article{meli2023planning,
  author  = {Meli, Daniele and Nakawala, Hirenkumar and Fiorini, Paolo},
  title   = {Logic programming for deliberative robotic task planning},
  journal = {Artificial Intelligence Review},
  volume  = {56},
  number  = {9},
  pages   = {9011--9049},
  year    = {2023},
  doi     = {10.1007/s10462-022-10389-w}
}

@article{filippov1964differential,
  title={Differential equations with discontinuous right-hand side},
  author={Filippov, Alexey F and others},
  journal={American Mathematical Society Translations},
  volume={42},
  number={2},
  pages={199--231},
  year={1964},
  publisher={Amer. Math. Soc.}
}

@article{scheinker2014non,
  title={Non-{C}2 {L}ie bracket averaging for nonsmooth extremum seekers},
  author={Scheinker, Alexander and Krsti{\'c}, Miroslav},
  journal={Journal of Dynamic Systems, Measurement, and Control},
  volume={136},
  number={1},
  pages={011010},
  year={2014},
  publisher={American Society of Mechanical Engineers}
}

@article{andrychowicz2017hindsight,
  title={Hindsight experience replay},
  author={Andrychowicz, Marcin and Wolski, Filip and Ray, Alex and Schneider, Jonas and Fong, Rachel and Welinder, Peter and McGrew, Bob and Tobin, Josh and Pieter Abbeel, OpenAI and Zaremba, Wojciech},
  journal={Advances in neural information processing systems},
  volume={30},
  year={2017}
}

@article{cho2024model,
  title={Model-based transfer learning for contextual reinforcement learning},
  author={Cho, Jung-Hoon and Jayawardana, Vindula and Li, Sirui and Wu, Cathy},
  journal={Advances in Neural Information Processing Systems},
  volume={37},
  pages={88279--88319},
  year={2024}
}

@inproceedings{liu2025locoformer,
  title={LocoFormer: Generalist Locomotion via Long-context Adaptation},
  author={Liu, Min and Pathak, Deepak and Agarwal, Ananye},
  booktitle={Conference on Robot Learning},
  pages={532--546},
  year={2025},
  organization={PMLR}
}

@inproceedings{saxena2026improved,
  title={Improved robustness of deep reinforcement learning for control of time-varying systems by bounded extremum seeking},
  author={Saxena, Shaifalee and Williams, Alan and Fierro, Rafael and Scheinker, Alexander},
  booktitle={2026 American Control Conference (ACC)},
  pages={1165--1172},
  year={2026},
  organization={IEEE}
}

@article{saxena2026mahalanobis,
  title={Mahalanobis-Guided Latent OOD Detection for Hybrid ES-DRL Control in Time-Varying Systems},
  author={Saxena, Shaifalee and Scheinker, Alexander},
  journal={arXiv preprint arXiv:2606.11474},
  year={2026}
}

@article{plappert2018multi,
  title={Multi-goal reinforcement learning: Challenging robotics environments and request for research},
  author={Plappert, Matthias and Andrychowicz, Marcin and Ray, Alex and McGrew, Bob and Baker, Bowen and Powell, Glenn and Schneider, Jonas and Tobin, Josh and Chociej, Maciek and Welinder, Peter and others},
  journal={arXiv preprint arXiv:1802.09464},
  year={2018}
}

@inproceedings{zeng2018learning,
  title={Learning synergies between pushing and grasping with self-supervised deep reinforcement learning},
  author={Zeng, Andy and Song, Shuran and Welker, Stefan and Lee, Johnny and Rodriguez, Alberto and Funkhouser, Thomas},
  booktitle={2018 IEEE/RSJ International Conference on Intelligent Robots and Systems (IROS)},
  pages={4238--4245},
  year={2018},
  organization={IEEE}
}

@inproceedings{deng2019deep,
  title={Deep reinforcement learning for robotic pushing and picking in cluttered environment},
  author={Deng, Yuhong and Guo, Xiaofeng and Wei, Yixuan and Lu, Kai and Fang, Bin and Guo, Di and Liu, Huaping and Sun, Fuchun},
  booktitle={2019 IEEE/RSJ International Conference on Intelligent Robots and Systems (IROS)},
  pages={619--626},
  year={2019},
  organization={IEEE}
}

@article{lauri2022partially,
  title={Partially observable {M}arkov decision processes in robotics: A survey},
  author={Lauri, Mikko and Hsu, David and Pajarinen, Joni},
  journal={IEEE Transactions on Robotics},
  volume={39},
  number={1},
  pages={21--40},
  year={2022},
  publisher={IEEE}
}

@inproceedings{cobbe2019quantifying,
  title={Quantifying generalization in reinforcement learning},
  author={Cobbe, Karl and Klimov, Oleg and Hesse, Chris and Kim, Taehoon and Schulman, John},
  booktitle={International Conference on Machine Learning},
  pages={1282--1289},
  year={2019},
  organization={PMLR}
}

@article{duan2016rl,
  title={{RL}$^{2}$: Fast reinforcement learning via slow reinforcement learning},
  author={Duan, Yan and Schulman, John and Chen, Xi and Bartlett, Peter L and Sutskever, Ilya and Abbeel, Pieter},
  journal={arXiv preprint arXiv:1611.02779},
  year={2016}
}

@inproceedings{finn2017maml,
  author    = {Chelsea Finn and Pieter Abbeel and Sergey Levine},
  title     = {Model-Agnostic Meta-Learning for Fast Adaptation of Deep Networks},
  booktitle = {Proceedings of the 34th International Conference on Machine Learning},
  series    = {Proceedings of Machine Learning Research},
  volume    = {70},
  pages     = {1126--1135},
  year      = {2017}
}

@inproceedings{benjamins2021carl,
  author    = {Carolin Benjamins and Theresa Eimer and Frederik Schubert and Andr{\'e} Biedenkapp and Bodo Rosenhahn and Frank Hutter and Marius Lindauer},
  title     = {{CARL}: A Benchmark for Contextual and Adaptive Reinforcement Learning},
  booktitle = {NeurIPS 2021 Workshop on Ecological Theory of Reinforcement Learning},
  year      = {2021},
  doi       = {10.48550/arXiv.2110.02102}
}

@inproceedings{johannink2019residual,
  title={Residual reinforcement learning for robot control},
  author={Johannink, Tobias and Bahl, Shikhar and Nair, Ashvin and Luo, Jianlan and Kumar, Avinash and Loskyll, Matthias and Ojea, Juan Aparicio and Solowjow, Eugen and Levine, Sergey},
  booktitle={2019 International Conference on Robotics and Automation (ICRA)},
  pages={6023--6029},
  year={2019},
  organization={IEEE}
}

@article{kulkarni2022learning,
  title={Learning assembly tasks in a few minutes by combining impedance control and residual recurrent reinforcement learning},
  author={Kulkarni, Padmaja and Kober, Jens and Babu{\v{s}}ka, Robert and Della Santina, Cosimo},
  journal={Advanced Intelligent Systems},
  volume={4},
  number={1},
  pages={2100095},
  year={2022},
  publisher={Wiley Online Library}
}

@inproceedings{scheinker2013model,
  title={Model independent beam tuning},
  author={Scheinker, Alexander and others},
  booktitle={Proceedings of the 2013 International Particle Accelerator Conference, Shanghai, China},
  year={2013},
  url={https://proceedings.jacow.org/IPAC2013/papers/tupwa068.pdf}
}

@article{scheinker2014extremum,
  title={Extremum seeking with bounded update rates},
  author={Scheinker, Alexander and Krsti{\'c}, Miroslav},
  journal={Systems \& Control Letters},
  volume={63},
  pages={25--31},
  year={2014},
  publisher={Elsevier},
  url={https://doi.org/10.1016/j.sysconle.2013.10.004}
}

@article{scheinker2016bounded,
  title={Bounded extremum seeking with discontinuous dithers},
  author={Scheinker, Alexander and Scheinker, David},
  journal={Automatica},
  volume={69},
  pages={250--257},
  year={2016},
  publisher={Elsevier},
  url={https://doi.org/10.1016/j.automatica.2016.02.023}
}

@inproceedings{silver2014deterministic,
  title={Deterministic policy gradient algorithms},
  author={Silver, David and Lever, Guy and Heess, Nicolas and Degris, Thomas and Wierstra, Daan and Riedmiller, Martin},
  booktitle={International Conference on Machine Learning},
  pages={387--395},
  year={2014},
  organization={PMLR}
}

@article{lillicrap2020continuous,
  title={Continuous control with deep reinforcement learning},
  author={Lillicrap, Timothy P and Hunt, Jonathan J and Pritzel, Alexander and Heess, Nicolas and Erez, Tom and Tassa, Yuval and Silver, David and Wierstra, Daan},
  journal={arXiv preprint arXiv:1509.02971},
  year={2015}
}

@inproceedings{guha2021online,
  title={Online policies for real-time control using MRAC-RL},
  author={Guha, Anubhav and Annaswamy, Anuradha M},
  booktitle={2021 60th IEEE Conference on Decision and Control (CDC)},
  pages={1808--1813},
  year={2021},
  organization={IEEE}
}

@article{romero2025actor,
  title={Actor--Critic Model Predictive Control: Differentiable Optimization Meets Reinforcement Learning for Agile Flight},
  author={Romero, Angel and Aljalbout, Elie and Song, Yunlong and Scaramuzza, Davide},
  journal={IEEE Transactions on Robotics},
  volume={42},
  pages={673--692},
  year={2025},
  publisher={IEEE}
}

@article{emam2021safe,
  title={Safe model-based reinforcement learning using robust control barrier functions},
  author={Emam, Yousef and Glotfelter, Paul and Kira, Zsolt and Egerstedt, Magnus},
  journal={arXiv preprint arXiv:2110.05415},
  year={2021}
}

@article{koropouli2016extremum,
  title={An extremum-seeking control approach for constrained robotic motion tasks},
  author={Koropouli, Vasiliki and Gusrialdi, Azwirman and Hirche, Sandra and Lee, Dongheui},
  journal={Control Engineering Practice},
  volume={52},
  pages={1--14},
  year={2016},
  publisher={Elsevier}
}

@inproceedings{xie2005autopilot,
  title={Autopilot-based nonlinear {UAV} formation controller with extremum-seeking},
  author={Xie, Feng and Zhang, Ximing and Fierro, Rafael and Motter, Mark},
  booktitle={Proceedings of the 44th IEEE Conference on Decision and Control},
  year={2005},
  organization={IEEE}
}

@article{scheinker2021extremum,
  title={Extremum seeking-based control system for particle accelerator beam loss minimization},
  author={Scheinker, Alexander and Huang, En-Chuan and Taylor, Charles},
  journal={IEEE Transactions on Control Systems Technology},
  volume={30},
  number={5},
  year={2021},
  publisher={IEEE},
  doi={10.1109/TCST.2021.3136133}
}

@mastersthesis{bajpai2024investigating,
  title={Investigating the performance of different controllers in optimized path tracking in robotics: A lie bracket system and extremum seeking approach},
  author={Bajpai, Shivam},
  year={2024},
  school={University of Cincinnati}
}

@inproceedings{de2022event,
  title={Event-driven adaptive Vertical Stabilization in tokamaks based on a bounded Extremum Seeking algorithm},
  author={De Tommasi, Gianmaria and Dubbioso, Sara and Mele, Adriano and Pironti, Alfredo},
  booktitle={2022 IEEE Conference on Control Technology and Applications (CCTA)},
  pages={831--836},
  year={2022},
  organization={IEEE}
}

@inproceedings{nguyen2019review,
  title={Review of deep reinforcement learning for robot manipulation},
  author={Nguyen, Hai and La, Hung},
  booktitle={2019 Third IEEE International conference on Robotic Computing (IRC)},
  pages={590--595},
  year={2019},
  organization={IEEE}
}

@article{billard2019trends,
  title={Trends and challenges in robot manipulation},
  author={Billard, Aude and Kragic, Danica},
  journal={Science},
  volume={364},
  number={6446},
  pages={eaat8414},
  year={2019},
  publisher={American Association for the Advancement of Science}
}

@inproceedings{todorov2012mujoco,
  title={{MuJoCo}: A physics engine for model-based control},
  author={Todorov, Emanuel and Erez, Tom and Tassa, Yuval},
  booktitle={2012 IEEE/RSJ International Conference on Intelligent Robots and Systems},
  pages={5026--5033},
  year={2012},
  organization={IEEE}
}

@article{zhou2017fast,
  title={A fast stochastic contact model for planar pushing and grasping: Theory and experimental validation},
  author={Zhou, Jiaji and Bagnell, J Andrew and Mason, Matthew T},
  journal={arXiv preprint arXiv:1705.10664},
  year={2017}
}

@inproceedings{akinola2021dynamic,
  title={Dynamic grasping with reachability and motion awareness},
  author={Akinola, Iretiayo and Xu, Jingxi and Song, Shuran and Allen, Peter K},
  booktitle={2021 IEEE/RSJ International Conference on Intelligent Robots and Systems (IROS)},
  pages={9422--9429},
  year={2021},
  organization={IEEE}
}
\nocite{}
\end{document}